\documentclass[10pt, a4paper]{article}

\usepackage[final]{lrec-coling2024} 

\usepackage{microtype}

\usepackage{times}
\usepackage{latexsym}
\usepackage{epsfig}
\usepackage{graphicx}

\usepackage{amsmath}
\usepackage{amssymb}
\usepackage{comment}
\usepackage{booktabs}
\usepackage{float}
\usepackage{multirow}
\usepackage{enumitem}
\usepackage{caption, subcaption, overpic, textpos}

\usepackage{xcolor}
\usepackage{colortbl}
\usepackage{makecell}
\definecolor{ours-highlight}{rgb}{0.86, 0.82, 1.0}
\definecolor{darkgray}{rgb}{0.66, 0.66, 0.66}

\title{Semantics-enhanced Cross-modal Masked Image Modeling \\ for Vision-Language Pre-training}

\name{
\begin{tabular}{c}
Haowei Liu$^{1,2*}$, Yaya Shi$^{3*}$, Haiyang Xu$^{4\dagger}$, Chunfeng Yuan$^{1\dagger}$, Qinghao Ye$^{4}$\\
Chenliang Li$^{4}$, Ming Yan$^{4}$, Ji Zhang$^{4}$, Fei Huang$^{4}$, Bing Li$^{1}$, Weiming Hu$^{1,2,5}$\\ 
\end{tabular}
} 

\address{
$^{1}$MAIS, Institute of Automation, Chinese Academy of Sciences, China \\
$^{2}$School of Artificial Intelligence, University of Chinese Academy of Sciences, China \\
$^{3}$University of Science and Technology of China
\ $^{4}$Alibaba Group \\
$^{5}$School of Information Science and Technology, ShanghaiTech University, China \\
liuhaowei2019@ia.ac.cn, shiyaya@mail.ustc.edu.cn \\
\{shuofeng.xhy, ym119608\}@alibaba-inc.com, \{cfyuan, bli, wmhu\}@nlpr.ia.ac.cn \\
}

\abstract{
In vision-language pre-training (VLP), masked image modeling (MIM) has recently been introduced for fine-grained cross-modal alignment. However, in most existing methods, the reconstruction targets for MIM lack high-level semantics, and text is not sufficiently involved in masked modeling. These two drawbacks limit the effect of MIM in facilitating cross-modal semantic alignment. In this work, we propose a semantics-enhanced cross-modal MIM framework (SemMIM) for vision-language representation learning. Specifically, to provide more semantically meaningful supervision for MIM, we propose a local semantics enhancing approach, which harvest high-level semantics from global image features via self-supervised agreement learning and transfer them to local patch encodings by sharing the encoding space. Moreover, to achieve deep involvement of text during the entire MIM process, we propose a text-guided masking strategy and devise an efficient way of injecting textual information in both masked modeling and reconstruction target acquisition. Experimental results validate that our method improves the effectiveness of the MIM task in facilitating cross-modal semantic alignment. Compared to previous VLP models with similar model size and data scale, our SemMIM model achieves state-of-the-art or competitive performance on multiple downstream vision-language tasks.
 \\ \newline  }

\begin{document}

\maketitleabstract

\section{Introduction}
\label{sec:intro}

\let\thefootnote\relax\footnotetext{$^*$Equal contribution.

\ \ \ $^\dagger$Corresponding authors.}

Learning transferable vision-language representations for various downstream multimodal tasks (\textit{e.g.} image-text retrieval, visual question answering and image captioning)
by pre-training on large-scale image-text datasets has attracted extensive attention in recent years~\cite{Tan2019LXMERTLC,Huang2020PixelBERTAI,li2021align,li2022mplug,ji2023seeing}.
For vision-language pre-training (VLP), the most crucial thing is to align the representation spaces of the two modalities.
To this end, contrastive learning between global image and text features is commonly adopted.
However, such coarse-grained alignment usually ignores detailed information in image and text.
To achieve finer-grained cross-modal alignment, recent work \cite{ge2022miles,he2022vlmae,Geng2022MultimodalMA,bao2022vl} has introduced the masked image modeling (MIM) task into VLP.

Despite the achievements of MIM in the vision domain, leveraging MIM to facilitate cross-modal semantic alignment and promote vision-language representation learning still faces significant challenges.
\textbf{1) Semantic issue.}
Different from text whose words are naturally discrete and abstract, the region information of image is continuous and ambiguous.
Therefore, without semantically meaningful supervision for MIM, the inconsistency in semantic levels between vision and language
will impair cross-modal alignment.
However, in most existing methods, the reconstruction targets provided for MIM lack high-level semantics.
For instance, VL-BEiT~\cite{bao2022vl} uses a discrete variational autoencoder (dVAE) to encode image patches into discrete codes as MIM's supervision.
As dVAE is trained through image reconstruction which simply minimizes pixel-level differences between original and reconstructed images, its patch encodings lack high-level visual semantics.
VLMAE~\cite{he2022vlmae} and M3AE~\cite{Geng2022MultimodalMA} take the raw pixels of the masked regions as the reconstruction targets of MIM, thus facing even more severe semantic issue.
\textbf{2) Insufficient text involvement.}
Most previous methods directly introduce MIM into VLP without elaborate designs for deep involvement of text. For example, in MILES \cite{ge2022miles} and MaskCLIP \cite{dong2023maskclip}, text features do not directly participate in the MIM process, which inevitably limits their contribution to cross-modal semantic interaction.
As for the selection of masked patches, most methods simply adopt a random masking strategy without considering the interaction between image and text.

To address these issues, we propose a semantics-enhanced cross-modal MIM framework (dubbed as \textbf{SemMIM}) to improve the effectiveness of MIM in facilitating vision-language semantic alignment.
In this framework, we first propose to inject high-level semantics into the local encodings of image patches via self-supervised agreement learning and sharing encoding space, which can thus provide more semantically meaningful supervision for MIM.
Moreover, we make elaborate designs for deep involvement of text during the MIM process (\textit{i.e.} masking strategy, masked modeling and reconstruction target acquisition) to further promote cross-modal interaction.
Specifically, we adopt a fusion architecture which includes an image encoder, a text encoder and a fusion encoder in the model.
For masked image modeling, we set an extra encoding head on top of the image encoder, which projects the masked visual tokens into an encoding space.

Due to the small granularity of image patches, it's challenging to directly acquire high-level semantics from them.
To tackle this, we propose to harvest high-level semantics from global visual features via self-supervised agreement learning, and transfer them into patch encodings by sharing the same encoding space.
Specifically, we derive a momentum image encoder from the original image encoder, whose parameters are obtained via the exponential moving average approach.
Similarly, a momentum encoding head is also derived on top of the momentum image encoder.
Given an input image, we apply random augmentation to obtain two different views of it, which are fed into the original and the momentum image encoders respectively.
Then we employ the encoding heads to map the obtained global visual features into the encoding space.
By learning the agreement between different views, we can acquire high-level visual semantics and shape the semantic structure of the encoding space.
In the meantime, we share the encoding space for global and local features, \textit{i.e.}, using the same momentum image encoder and encoding head to transform patch features into visual encodings.
The learned high-level semantics are thus transferred to the local patch encodings.
In this manner, we obtain semantics-enhanced local encodings and provide more semantically meaningful reconstruction targets for MIM.

Moreover, to further enhance MIM's ability of facilitating cross-modal semantic alignment, we make three elaborate designs for deep text involvement throughout the entire MIM process.
1) We leverage the semantic relevance between image patches and the paired text to select masked patches, which can encourage the utilization of textual information during masked modeling.
2) We devise an efficient way of fusing textual information while masked modeling.
This allows recovering the masked regions by reasoning both contextual visual information and paired textual information, and thus promotes cross-modal semantic interaction.
3) We fuse text features when transforming image patches into visual encodings with the momentum image encoder, so that high-level textual semantics can also be injected into the encodings.

Visualization results demonstrate that the proposed SemMIM framework can learn high-level visual semantics effectively.
And extensive experiments show our model achieves state-of-the-art or competitive performance on multiple downstream vision-language tasks, which validates the superiority of our method.

Our contributions can be summarized as follows:
\begin{itemize}
\item We propose a semantics-enhanced, text-deeply-involved MIM framework to facilitate fine-grained cross-modal semantic alignment in vision-language pre-training.
\item High-level visual semantics are injected into image patch encodings via self-supervised learning and shared encoding space, thus providing more semantically meaningful reconstruction targets for MIM.
\item Our elaborate designs for deep text involvement throughout the entire MIM process further promote cross-modal interaction, including masking strategy, masked modeling and reconstruction target acquisition.
\end{itemize}

\section{Related Work}

\subsection{Vision-Language Pre-training}

Large-scale vision-language pre-training \cite{Lu2019ViLBERTPT,Huang2020PixelBERTAI,Zhou2020UnifiedVP,li2021align,xu2021e2e,li2022mplug,xu2023mplug,shi2023learning} has shown promising performance on various downstream vision-language tasks.
To mine the associations between image and text, various pre-training objectives are proposed, which can be divided into two categories: discriminative tasks and generative tasks.
Two commonly-used discriminative tasks are image-text contrastive learning (ITC) and image-text matching (ITM).
ITC is used to align the global representations of image and text.
ITM aims to predict whether an image and a text match with each other in the cross-modal representation space. 
Generative tasks aim to reconstruct the corrupted text (or image) with the help of vision (or text) modality.
For instance, masked language modeling (MLM) predicts the masked tokens by reasoning contextual text information and the paired image information \cite{wu2022cross,chen2023duet}.
In this work, we focus on improving the effectiveness of the Masked Image Modeling (MIM) task, so as to better facilitate fine-grained cross-modal alignment.

\begin{figure*}[t]
\centering
\setlength{\abovecaptionskip}{1mm}
\includegraphics[width=0.98\linewidth]{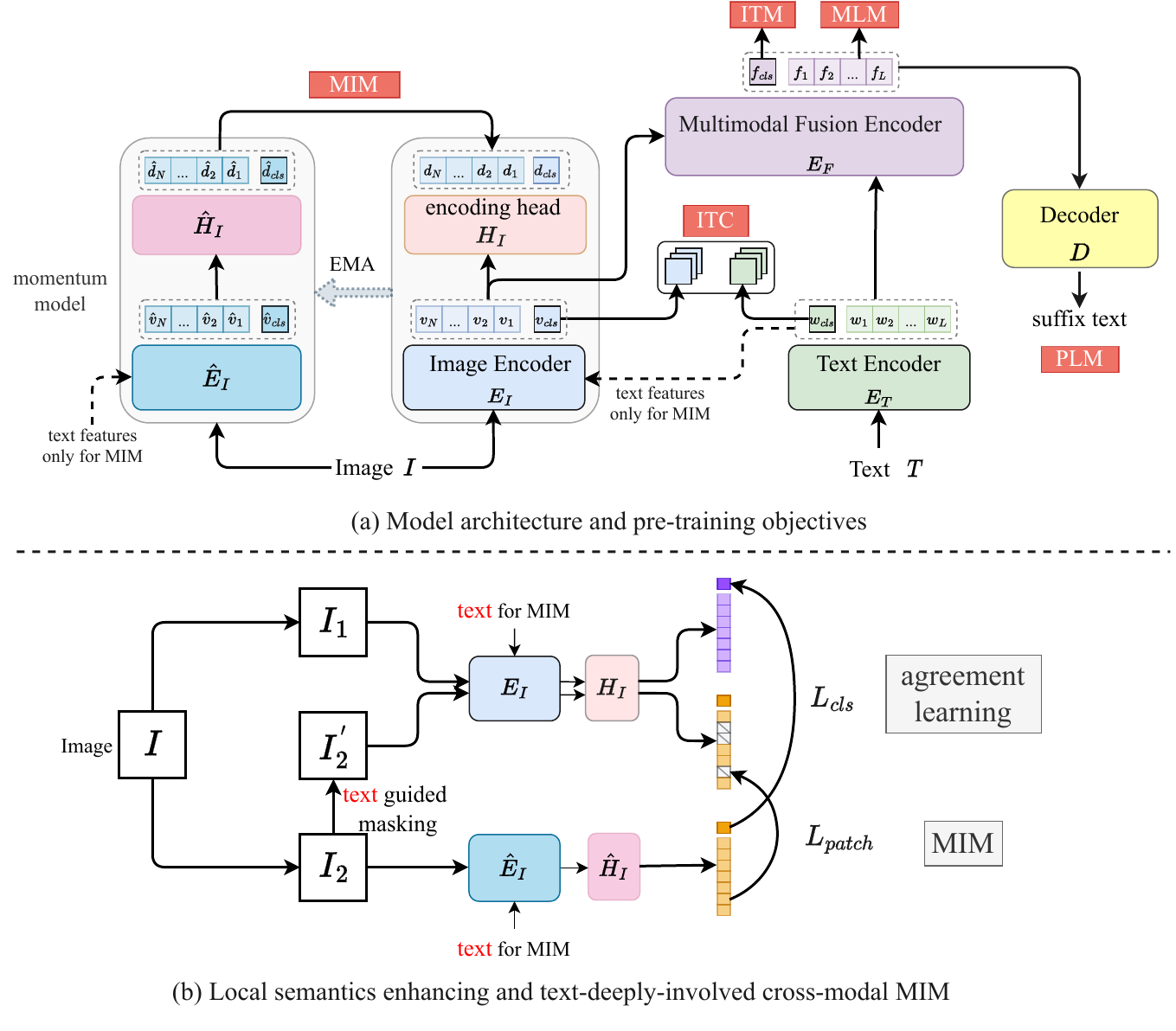}
\caption{
Overview of our method. (a) shows the model architecture and pre-training objectives of our SemMIM framework. (b) illustrates the proposed local semantics enhancing approach, which harvests high-level semantics from global visual features via agreement learning, and transfer them into patch encodings by sharing the same encoding space. And (b) also shows our designs for deep involvement of text during MIM, including text-guided masking strategy and injecting textual information into both masked modeling and reconstruction target acquisition.}
\label{fig: main}
\end{figure*}

\subsection{Masked Image Modeling in VLP}
Inspired by the success of MLM in multimodal learning, recent methods \cite{ge2022miles,he2022vlmae,Geng2022MultimodalMA,bao2022vl} have introduced MIM into VLP for fine-grained cross-modal alignment.
VLMAE~\cite{he2022vlmae} and M3AE~\cite{Geng2022MultimodalMA} follow MAE~\cite{he2022masked} to recover the raw pixels of the masked image regions with an autoencoder architecture.
Following BEiT \cite{bao2021beit}, VL-BEiT~\cite{bao2022vl} and BEiT-v3~\cite{Wang2022ImageAA} utilize a pre-trained discrete variational autoencoder (dVAE) to encode image patches into discrete codes as the supervision for MIM.
As dVAE is trained through image reconstruction which minimizes pixel-level differences between original and reconstructed images, its encodings concentrate more on low-level visual information.
Therefore, the above methods can't provide high-level semantic supervision for MIM.
Such inconsistency in semantic levels between vision and language impairs MIM's effectiveness in facilitating cross-modal alignment.
Another issue of existing methods is insufficient text involvement in MIM.
For instance, in MILES \cite{ge2022miles} and MaskCLIP \cite{dong2023maskclip}, text features do not directly participate in the MIM process.
Thus during masked modeling, the model only reasons visual information to recover masked regions, without direct vision-language interaction.
This inevitably limits MIM's benefit to cross-modal semantic alignment.
In this work, we propose a semantics-enhanced cross-modal MIM framework, which tackles the above stated issues by local semantics enhancing and text-deeply-involved cross-modal MIM.

\section{Method}

\subsection{Model Architecture}
\label{sec: model arch}

As shown in Figure~\ref{fig: main}(a), our SemMIM framework contains an image encoder $E_{I}$(·), a text encoder $E_{T}$(·), a multi-modal fusion encoder $E_{F}$(·), a decoder $D$(·) and an additional momentum image encoder $\hat{E}_{I}$(·). 
All the encoders are Transformer models.
The image encoder splits the input image into patches and encodes them as a sequence of visual features 
$\left\{v_{cls}, v_1, v_2, \ldots, v_N\right\}$.
The text encoder tokenizes the input text into tokens and projects them into a sequence of word features $\left\{w_{cls}, w_1, w_2, \ldots, w_L\right\}$.
$v_{cls}$ and $w_{cls}$ are the features of the [CLS] token for image and text respectively.
$N$ is the number of image patches, and $L$ is the number of text tokens.
The multimodal fusion encoder fuses the visual features and the word features through cross attention.
The obtained cross-modal representations are fed into the decoder for sequence to sequence learning,
which equips our model with both understanding and generation abilities.

\subsection{Semantics-enhanced Cross-modal Masked Image Modeling}

\subsubsection{Local Semantics Enhancing}
\label{sec: sec3.2.1}

As pointed in PeCo \cite{dong2023peco} and BEiT v2 \cite{peng2022beit}, a significant drawback of the offline visual tokenizers (\textit{e.g.} dVAE) adopted by previous methods is that their encodings lack high-level semantics.
Instead, they concentrate more on low-level visual information such as color and texture.
Taking such visual encodings as supervision inevitably hinders MIM’s objective of cross-modal semantic alignment.
However, as the granularity of image patches is rather small, it's challenging to directly acquire high-level semantics from them.
Therefore, we propose to harvest high-level semantics from global image features via self-supervised learning, which are further transferred to local patch features by sharing the same encoding space.

Unlike using pre-trained dVAE or raw pixels as reconstruction targets, we adopt an online approach and utilize the model itself to provide supervision for MIM.
As shown in Figure~\ref{fig: main}(a), we derive a momentum image encoder $\hat{E}_I$ from the original image encoder $E_I$.
$\hat{E}_I$ shares the same structure with $E_I$, and its parameters are obtained via exponential moving averaging the parameters of $E_I$ during the iterative updating process of the model:
\begin{equation}
\theta_{\hat{E}_{I}}=m \theta_{\hat{E}_{I}}+(1-m) \theta_{E_{I}},
\end{equation}
where $m$ is a momentum coefficient.
Besides, we add two encoding heads $H_I$ and $\hat{H}_I$ on top of the original and the momentum image encoders respectively.
Each encoding head is an MLP module, which transforms the extracted image features into categorical distributions in a $K$-dimensional encoding space.
Here the parameters of $\hat{H}_I$ are also obtained via exponential moving averaging the parameters of $H_I$.
Previous methods such as dVAE encode image patches into discrete codes, \textit{i.e.}, their categorical distributions are one-hot.
Instead, to preserve more information, we adopt a softmax form of categorical distributions. In other words, the reconstruction target of each masked image patch is a soft label.

To obtain more semantically meaningful visual encodings, we first harvest high-level semantic information from global image features.
Specifically, as illustrated in Figure \ref{fig: main}(b), given an input image $I$, we first generate two different distorted views of it $I_1$ and $I_2$ through random augmentation.
The two views are then fed into the momentum and original image encoders respectively, and get the corresponding encodings of their [CLS] tokens ${d}_1^{cls}$ and $\hat{d}_2^{cls}$.
Inspired by DINO \cite{caron2021emerging}, we learn the agreement between view 1 and view 2 as follows:
\begin{equation}
\mathcal{L}_{cls}=\mathbb{E}(\mathrm{CE}({d}_{1}^{cls}, \hat{d}_{2}^{cls})),
\label{equ: ssl}
\end{equation}
where $\mathrm{CE}$ denotes cross entropy loss.
Through this bootstrapping process on the [CLS] tokens, we acquire high-level semantics from the global image features and shape the semantic structure of the encoding space.
In the meantime, we use the same momentum image encoder and encoding head to transform local patch features into visual encodings, \textit{i.e.}, the [CLS] tokens and the patch tokens share the same encoding space.
Therefore, the learned high-level semantics can be naturally transferred to the local patch encodings.
In this way, we achieve local semantics enhancing and can provide more semantically meaningful reconstruction targets for the MIM task.

\subsubsection{Text-deeply-involved Cross-modal MIM}

By enhancing local semantics, we provide more semantically meaningful reconstruction targets for MIM.
However, lacking text involvement would lead MIM to degrade into learning purely visual information, which hinders facilitating cross-modal semantic alignment.
Therefore, we further make three elaborate designs within our framework for deep involvement of text during the entire MIM process.

Firstly, we propose an text-guided masking strategy.
Instead of masking a proportion of image patches randomly, we leverage semantic similarities between image patches and text for masked patch selection, so as to encourage the utilization of textual information during MIM.
Specifically, we first compute the similarity $S_i$ between each local patch feature and the global feature of the paired text.
Then we normalize the similarity scores of all patches using softmax, and take them as sampling probabilities to select a predefined proportion of patches.
In this way, image regions with higher relevance to text are more likely to be selected, thereby encouraging the model to recover masked regions by reasoning textual information.

Secondly, we devise an efficient way of fusing textual information while masked modeling, which enables the model to recover masked regions by reasoning both contextual visual information and paired textual information.
In cross-modal MLM, an extra fusion encoder is often required.
Instead, we concatenate the output of the text encoder with visual tokens at the last $N$ layers of the image encoder.
Thus the masked tokens can interact with the paired text features via self-attention.
We also investigate the effect of the start visual layer used to inject textual information for MIM. Please refer to the ablation study in Section \ref{sec: ablation}.

Thirdly, we also fuse text features when using the momentum encoder to transform image patches into encodings, which injects high-level textual semantics into patch encodings.
Note that the bootstrapping on [CLS] tokens in Section \ref{sec: sec3.2.1} would fail if textual information is leaked into [CLS] tokens.
Therefore, we infer twice at the last $N$ layers of the momentum encoder $\hat{E}_I$.
One for regular visual modeling to extract [CLS] tokens, the other for injecting textual information into patch encodings.

In Equation \ref{equ: ssl}, we feed the two views $I_1$ and $I_2$ of the input image into the momentum and original image encoders respectively.
Here, to perform cross-modal MIM, we first apply the proposed masking strategy to replace the selected image patches with a learnable [MASK] token.
Then we input the corrupted image ${I}_2^{'}$ into the image encoder $E_I$ to recover the information of the masked regions.
We denote $d_{2,i}^{patch}$ as the categorical distribution of the $i$-th masked patch of ${I}_2^{'}$, which is predicted by the encoding head $H_I$.
And $\hat{d}_{2,i}^{patch}$ for that generated by the momentum encoding head $\hat{H}_I$.
The loss of the MIM task can be formulated as follows:
\begin{equation}
\mathcal{L}_{patch}=\mathbb{E}(\frac{1}{M}\sum_{i=1}^M \mathrm{CE}(d_{2, i}^{patch}, \hat{d}_{2, i}^{patch })),
\end{equation}
where $\mathrm{CE}$ denotes cross-entropy loss. $M$ is the number of masked patches.

\subsubsection{Pre-training Objectives}
\label{sec: obj}

In addition to the cross-modal MIM task, we also jointly train the pre-training tasks below to learn cross-modal alignment following existing methods \cite{li2021align,li2022mplug,xu2023mplug}:

\noindent\textbf{Image-Text Contrastive Learning} (ITC, $\mathcal{L}_{\mathrm{itc}}$)
aims to learn better uni-modal representations before fusion.
\noindent\textbf{Image-Text Matching} (ITM, $\mathcal{L}_{\mathrm{itm}}$)
uses a binary classification task to predict whether a pair of image and text is matched with each other. 
\noindent\textbf{Masked Language Modeling} (MLM, $\mathcal{L}_{\mathrm{mlm}}$)
aims to predict the randomly masked words based on cross-modal representations.
\noindent\textbf{Prefix Language Modeling} (PLM, $\mathcal{L}_{\mathrm{plm}}$)
aims to equip the model with generation capability.
Given an image and the prefix text $T^{p}$, we use the decoder to predict the suffix text $T^{s}$. 
Following previous methods \cite{li2021align,li2022mplug}, we assign equal loss weights to all these pre-training objectives, and thus the full pre-training objective is:
\begin{equation}
\mathcal{L} =  \mathcal{L}_{\mathrm{cls}} + \mathcal{L}_{\mathrm{patch}} + \mathcal{L}_{\mathrm{itc}} + \mathcal{L}_{\mathrm{itm}} + \mathcal{L}_{\mathrm{mlm}} + 
\mathcal{L}_{\mathrm{plm}}
\end{equation}

\begin{table*}[t]
\setlength{\tabcolsep}{1.6mm}
\centering
\scalebox{0.75}{
\begin{tabular}{l|c|cccccc|cccccc}
\toprule[1.0pt]
\multicolumn{1}{l|}{\multirow{2}{*}{Methods}}      &
\multicolumn{1}{c|}{\# Pretrain} &
\multicolumn{6}{c|}{MSCOCO (5K test set)} & \multicolumn{6}{c}{Flickr30K (1K test set)} \\
      &  data & \multicolumn{3}{c}{TR} & \multicolumn{3}{c|}{IR} & \multicolumn{3}{c}{TR} & \multicolumn{3}{c}{IR}          \\
\midrule
&&R@1&R@5&R@10&R@1&R@5&R@10&R@1&R@5&R@10&R@1&R@5&R@10 \\
\midrule
UNITER \cite{chen2020uniter} & 4M     & 65.7&88.6&93.8&52.9&79.9&88.0&87.3& 98.0&99.2&75.6&94.1&96.8  \\
OSCAR \cite{Li2020OscarOA} & 4M  & 70.0&91.1&95.5&54.0&80.8&88.5&-& -&-&-&-&-   \\
ALBEF \cite{li2021align} & 4M & 73.1 &91.4 &96.0 &56.8 &81.5 &89.2 &94.3 &99.4 &99.8 &82.8 &96.7 &98.4 \\
VinVL \cite{zhang2021vinvl}  &4M      &   74.6                     & 92.6                  & 96.3                  & 58.1 & 83.2  & 90.1                  &   -                  &            -           &      -                &                 -       &          -              &         -              \\
ALIGN \cite{jia2021scaling} & 1.8B  & 77.0&93.5&96.9&59.9&83.3&89.8&95.3& 99.8&100.0&84.9&97.4&98.6   \\
VLMAE \cite{he2022vlmae} & 4M &77.3 &93.6 & 97.4 &59.6 &83.6 &90.3 &95.2 &99.6 &99.9 &83.6 &96.6 &98.5 \\
SCL \cite{ji2023seeing} & 4M & 77.7 & 94.1 & 97.4 & 60.1 & 84.5 & 91.4 & 95.9 & 99.8 & 100.0 & 84.5 & 97.4 & 98.9 \\
MAP \cite{ji2023map} & 4M & 79.3 & 94.8 & 97.6 & 60.9 & 86.2 & 93.1 & 94.9 & 99.5 & 99.8 & 83.8 & 97.2 & 98.7 \\
VL-BEiT \cite{bao2022vl} & 4M & 79.5 &-&- &61.5 &- &- &95.8 &-&- &83.9 &- &- \\
BLIP \cite{li2022blip} & 14M & 80.6 &95.2&97.6&63.1&85.3&91.1&96.6& 99.8&100.0&87.2&97.5&98.8                 \\
X-VLM \cite{zeng2021multi} & 4M & 80.4 & 95.5&98.2 & 63.1 & 85.7 &91.6 & 96.8 & 99.8 & 100.0 & 86.1 & 97.4 &98.7\\
mPLUG \cite{li2022mplug} & 4M  & 80.5 &95.4&97.9&63.3&85.3&91.2&96.7& 99.8&100.0& 86.5&97.5&98.8   \\
\midrule
Ours & 4M &    \textbf{81.5}                   &        \textbf{ 96.2}                 &     \textbf{98.3}                     &       \textbf{64.2}                  &                \textbf{86.1}         &           \textbf{91.6} & \textbf{97.0}                    &               \textbf{99.8}          &      \textbf{100.0}                   &         \textbf{86.9}                 &      \textbf{97.8}                    &           \textbf{99.0}                         \\

\bottomrule[1.0pt]
\end{tabular}}          \\
\caption{Evaluation results of image-text retrieval on MSCOCO and Flickr30K datasets.}
\label{table:retrieval}
\end{table*}

\begin{table*}[t]
\setlength\tabcolsep{4pt}
\centering
\scalebox{0.75}{
\begin{tabular}{l|c|cccccccc|cc}
\toprule[1.0pt]
\multicolumn{1}{l|}{\multirow{3}{*}{Methods}}      &
\multicolumn{1}{c|}{\# Pretrain} &
\multicolumn{8}{c|}{COCO Caption} & \multicolumn{2}{c}{\multirow{1}{*}{NoCaps}}  \\
\multicolumn{1}{c|}{\multirow{2}{*}{}}      &
\multicolumn{1}{c|}{\multirow{2}{*}{Data}} &
\multicolumn{4}{c}{Cross-entropy Optimization} & \multicolumn{4}{c|}{CIDEr Optimization} & \multicolumn{2}{c}{zero-shot}  \\
      &   & B@4 & M & C & S & B@4 & M & C & S & C & S     \\
      
\midrule      
OSCAR \cite{Li2020OscarOA} & 6.5M  & - & - & - & - & 41.7 & 30.6 & 140.0 & 24.5 & 83.4 & 11.4 \\
VinVL \cite{zhang2021vinvl} & 5.65M  & 38.5 & 30.4 & 130.8 & 23.4 & 41.0 & 31.1 & 140.9 & 25.2 & 97.3 & 13.8 \\
BLIP \cite{li2022blip}  & 14M  & 38.6 & - & 129.7 & - & - & - & - & - & 105.1 & 14.4  \\
SimVLM$_{base}$ \cite{wang2021simvlm} & 1.8B  & 39.0 & 32.9 & 134.8 & 24.0 & - & - & - & - & - & - \\
mPLUG \cite{li2022mplug} & 4M & 39.3 & 30.1 & 132.4 & 23.3 & 41.2 & 30.8 & 140.2 & 25.2  & 98.3 & 12.9 \\
\midrule

Ours & 4M & 39.8 & 30.9 & 133.5 & 23.9 & 41.7 & 31.3 & 141.5 & 25.4  & 98.8 & 13.5  \\
\bottomrule[1.0pt]
\end{tabular}}
\caption{Evaluation results on COCO Caption ``Karpathy'' test split and NoCaps validation set. B@4: BLEU@4, M: METEOR, C: CIDEr, S: SPICE.} 
\label{table:caption}
\end{table*}

\begin{table}[t]
\setlength\tabcolsep{4pt}
\centering
\scalebox{0.75}{
\begin{tabular}{l|c|cc}
\toprule[1.0pt]
\multicolumn{1}{l|}{\multirow{2}{*}{Methods}}      &
\multicolumn{1}{c|}{\# Pretrain} &
\multicolumn{2}{c}{VQA} \\
\multicolumn{1}{l|}{ } & \multicolumn{1}{c|}{Data} & Test-dev  & Test-std \\

\midrule      
E2E-VLP \cite{xu2021e2e} & 4M & 73.25  & 73.67 \\
OSCAR \cite{Li2020OscarOA} & 6.5M & 73.16 & 73.44  \\
ALBEF \cite{li2021align} & 4M & 74.54 & 74.70  \\
VLMAE \cite{he2022vlmae} & 4M & 75.3 &75.4 \\
PTP-BLIP \cite{wang2023position} & 4M & 75.5 & 75.9 \\
ALBEF \cite{li2021align} & 14M & 75.84 & 76.04  \\
VinVL \cite{zhang2021vinvl} & 5.65M & 76.52 & 76.60 \\
VL-BEiT \cite{bao2022vl} & 4M & 77.53 & 77.75 \\
BLIP \cite{li2022blip}  & 14M & 77.54 & 77.62  \\
METER \cite{dou2022empirical} & 4M & 77.68 & 77.64   \\
mPLUG \cite{li2022mplug} & 4M & 77.58  & 77.73 \\
FLM$_{large}$ \cite{wang2023accelerating} & 4M & 77.80 & 77.84 \\
SimVLM$_{base}$ \cite{wang2021simvlm} & 1.8B & 77.87 & 78.14  \\

\midrule

Ours & 4M & \textbf{78.12}  & \textbf{78.18}  \\
\bottomrule[1.0pt]
\end{tabular}}
\caption{Evaluation results on VQA.} 
\label{table:vqa}
\end{table}

\section{Experiments}

\subsection{Pre-training Datasets}
Following ALBEF \cite{li2021align}, we use a hybrid pre-training dataset with 4 million images and 5.1 million captions,
including two in-domain datasets (MS COCO \cite{Lin2014MicrosoftCC}, Visual Genome \cite{krishna2017visual}), and two web-crawled datasets (Conceptual Captions \cite{sharma2018conceptual}, SBU Captions \cite{ordonez2011im2text}).

\subsection{Implementation Detail}

We adopt ViT-B/16 as the visual encoder, and initialize it with CLIP-ViT \cite{radford2021learning} which is pre-trained on 400 million noisy image-text pairs.
Besides, we use a 6-layer Transformer architecture for both the text encoder and the fusion encoder, and a 12-layer Transformer architecture for the decoder.
The text encoder and the fusion encoder are initialized using the first 6 layers and the last 6 layers of the $\text{BERT}_\text{base}$ model respectively.
For pre-training, we use the AdamW \cite{loshchilov2017decoupled} optimizer with a weight decay of 0.02.
We use different learning rates for the visual encoder and the rest parts of our model, which are warmed-up to 1e-5 (ViT-B/16) and 1e-4 ($\text{BERT}_\text{base}$) in the first 1000 iterations, and decay to 1e-6 following a cosine schedule.
We pre-train our model for 20 epochs on 8 NVIDIA A100 GPUs, with a total batch size of 1024.
The image resolution is 224$\times$224 for pre-training and is increased to 384$\times$384 for fine-tuning.
For the encoding head, we set the dimension of the categorical distribution (\textit{i.e.} the $K$ value) to 1024.

\subsection{Evaluation on Downstream Vision-Language Tasks}

\textbf{Image-Text Retrieval.}
We conduct experiments on MSCOCO and Flickr30K \cite{young2014image} datasets and adopt a two-stage retrieval strategy following ALBEF \cite{li2021align}, which takes a fast retrieval via uni-modal encoders and then reranks via the fusion encoder.
As shown in Table 1, our model outperforms previous methods on both datasets.
Note that VLMAE \cite{he2022vlmae} and VL-BEiT \cite{bao2022vl} are also MIM-based methods.
VL-BEiT adopts an offline tokenizer (\textit{i.e.} dVAE) to provide reconstruction targets for MIM, while VLMAE directly recovers the raw pixels of the masked regions.
We outperform these two methods by a large margin on both MSCOCO and Flickr30K.
This shows that, by injecting high-level semantics into local patch encodings via self-supervised learning and shared encoding space, our model can provide more semantically meaningful supervision for MIM and improves the effectiveness of MIM for cross-modal alignment.
Besides, our model also surpasses BLIP \cite{li2022blip}, which uses a significantly bigger dataset of 14 million image-text pairs for pre-training.

\vspace{0.3em}
\noindent\textbf{Image Captioning.}
We evaluate the image caption generation ability of our model on COCO Caption~\cite{Lin2014MicrosoftCC} and NoCaps~\cite{agrawal2019nocaps} datasets.
We fine-tune our model on COCO Caption, and test on the ``Karpathy'' test split and NoCaps validation set.
Following OSCAR~\cite{Li2020OscarOA}, we first fine-tune our model with cross-entropy loss, and then with CIDEr optimization~\cite{scst} for 5 extra epochs.
As shown in Table~\ref{table:caption}, our method surpasses mPLUG \cite{li2022mplug} which uses the same amount of pre-training data.
On COCO Caption, our method even outperforms some methods which use much more data for pre-training.
For instance, we surpass BLIP which uses 14M image-text pairs by a large margin of 3.8 on the CIDEr metric.
Moreover, on the NoCaps dataset which reflects the zero-shot caption generation ability, we also outperform previous methods
\cite{Li2020OscarOA, zhang2021vinvl, li2022mplug} which use similar amounts of pre-training data, showing the good generalization ability of our method.

\vspace{0.3em}
\noindent\textbf{Visual Question Answering.}
We treat VQA as an answer-generation task and directly apply unconstrained open-vocabulary generation during inference.
Following OFA \cite{wang2022unifying}, we concatenate the question with the object labels and OCR tokens extracted from images.
As Table \ref{table:vqa} shows, our model outperforms previous methods using similar amounts of pre-training data.
And we even surpass $SimVLM_{base}$ which uses 400$\times$ more image-text data for pre-training.
We largely outperform VLMAE by 2.82\% on test-dev and 2.78\% on test-std, and surpass VL-BEiT by 0.6\% on the test-dev split.
The comparison validates that,
with the proposed local semantics enhancing and text-deeply-involved cross-modal MIM,
our SemMIM framework equips the model with more powerful semantic reasoning ability.

\begin{table}[t]
\centering
\resizebox{0.38\textwidth}{!}{
\begin{tabular}{c|ccc}
\toprule[1pt]
\multirow{2}{*}{Supervision for MIM} & \multicolumn{2}{c}{MSCOCO}  & VQA \\ \cmidrule(l){2-4} 
 & TR           & IR                & Test-dev     \\ 
\midrule
w/o MIM      & 79.0         & 62.3              & 77.20  \\ 
\midrule
raw pixels                         & 78.6         & 62.2              & 77.08  \\ 
\midrule
dVAE    & 79.8           & 62.7           & 77.54  \\ 
\midrule
Ours     & \textbf{81.5}           & \textbf{64.2}          & \textbf{78.12}  \\ 
\bottomrule[1pt]
\end{tabular}}
\caption{Performance comparison of different supervision for MIM. Retrieval results are reported in terms of the R@1 of TR and IR.}
\label{tab:tokenizer}
\end{table}

\begin{table}[t]
\centering
\scalebox{0.8}{
\begin{tabular}{c|ccc}
\toprule[1pt]
\multirow{2}{*}{Masking Strategy} & \multicolumn{2}{c}{MSCOCO}  & VQA \\ \cmidrule(l){2-4}
                & TR            & IR          & Test-dev            \\ \midrule
random           & 81.0          & 63.6        & 77.96               \\ \midrule
text-guided     & \textbf{81.5} & \textbf{64.2} & \textbf{78.12}  \\
\bottomrule[1pt]
\end{tabular}}
\caption{Ablation on the masking strategy. R@1 retrieval results are reported.}
\label{tab:masking strategy}
\end{table}

\begin{table}[t]
\centering
\scalebox{0.8}{
\begin{tabular}{c|ccc}
\toprule[1pt]
\multirow{2}{*}{Mask Ratio} & \multicolumn{2}{c}{MSCOCO}  & VQA \\ \cmidrule(l){2-4} 
                            & TR            & IR          & Test-dev                    \\ \midrule
15\%                        & 80.8          & 63.6        & 77.65               \\ \midrule
30\%                        & \textbf{81.5} & \textbf{64.2} & \textbf{78.12}     \\ \midrule
45\%                        & 80.6            & 63.4          & 77.71         \\  \bottomrule[1pt]
\end{tabular}}
\caption{Effect of the mask ratio. Retrieval results are reported in terms of the R@1 of TR and IR.}
\label{tab:mask ratio}
\end{table}

\begin{table}[t]
\centering
\scalebox{0.8}{
\begin{tabular}{c|ccc}
\toprule
\multirow{2}{*}{$i_{th}$} & \multicolumn{2}{c}{MSCOCO} & VQA \\ \cmidrule(l){2-4} 
                      & TR           & IR          & Test-dev                       \\ \midrule
-                     & 80.4         & 63.2        & 77.63                \\ \midrule
1                     & 80.0         & 62.9          & 77.35                    \\ \midrule
6                     & 80.9           & 63.6          & 77.90                    \\ \midrule
9                     & 81.5           & 63.9          & 78.06                      \\ \midrule
12                    & \textbf{81.5}           & \textbf{64.2}                    & \textbf{78.12}       \\ \bottomrule
\end{tabular}}
\caption{The effect of the start visual layer used to inject textual information for MIM. Results are reported in terms of the R@1 of TR and IR.}
\label{tab: visual layer}
\end{table}

\subsection{Ablation Study}
\label{sec: ablation}

\textbf{The effect of local semantics enhancing.}
As shown in Table~\ref{tab:tokenizer}, we compare different supervision for the MIM task.
Compared to the baseline model which doesn't perform MIM, in our experiment, using raw pixels as the reconstruction targets for MIM in fact brings a slight performance drop.
We analyze this is due to raw pixels don't have any semantic abstraction, and thus taking them as MIM's supervision can't achieve effective cross-modal semantic interaction.
Using a pre-trained dVAE to provide supervision improves performance.
However, the boost is relatively limited, as the dVAE encodings also lack high-level semantics.
In contrast, our method significantly surpasses raw-pixels-based and dVAE-based MIM, showing that our proposed local semantics enhancing approach can largely improve the effectiveness of MIM.

\vspace{0.3em}
\noindent\textbf{The effect of the text-guided masking strategy.}
We compare the performance of random masking and the proposed text-guided masking.
As Table \ref{tab:masking strategy} shows, applying text-guided masking brings significant performance boost.
As image regions with higher relevance to text are more likely to be selected, text-guided masking encourages the model to recover masked
regions by reasoning textual information, and thus further facilitates cross-modal semantic interaction and fusion.

\begin{figure*}[t]
\centering
\setlength{\abovecaptionskip}{1mm}
\includegraphics[width=0.84\linewidth]{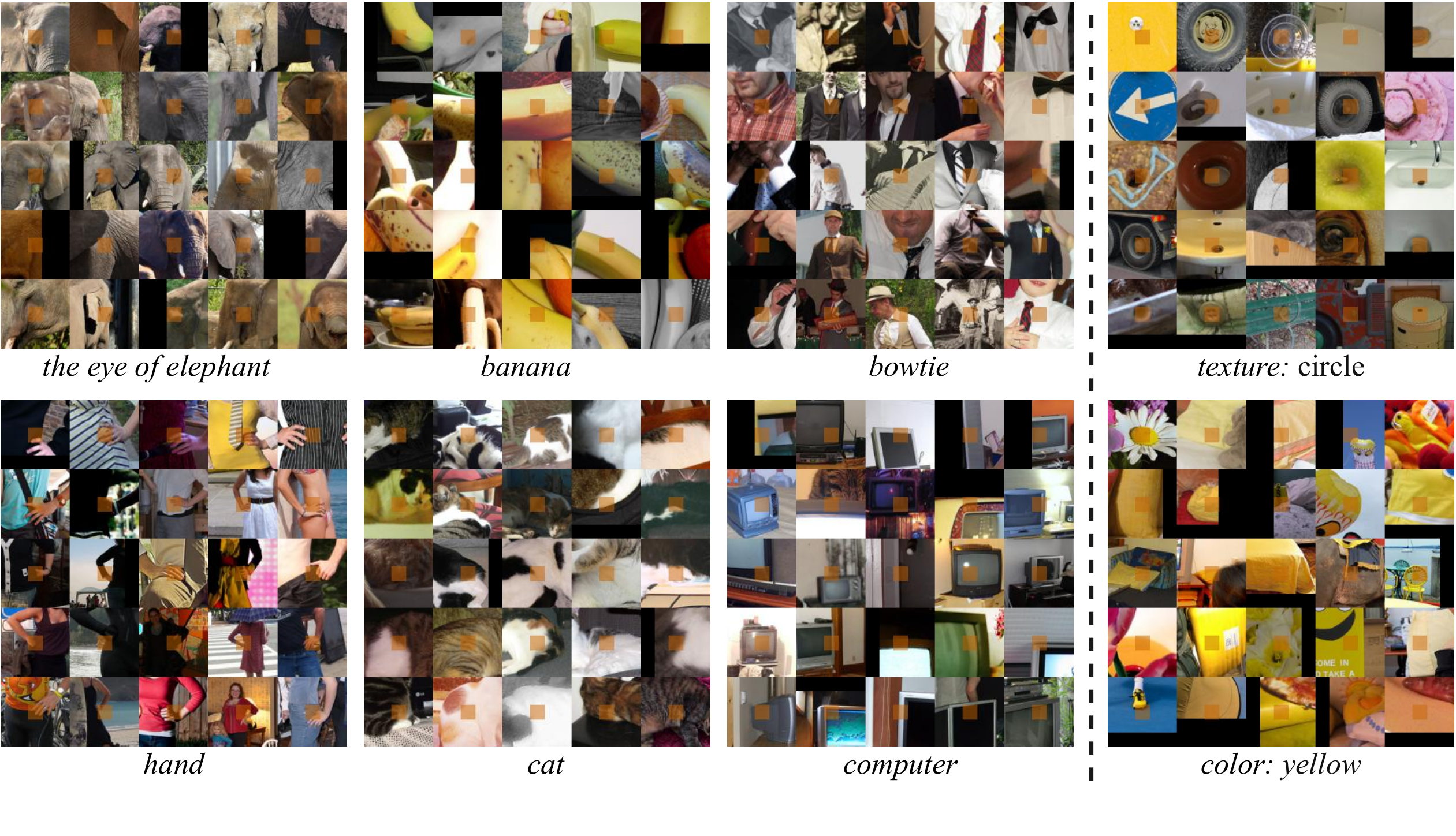}
\caption{Pattern clusters of image patch encodings. The left six figures (our model) showcase high-level semantic patterns \textit{the eye of elephant}, \textit{banana}, \textit{bowtie}, \textit{hand}, \textit{cat} and \textit{computer}. The right two figures (dVAE) showcase relatively low-level visual patterns \textit{circle texture} and \textit{yellow color}.}
\label{fig: pattern_cluster}
\end{figure*}

\begin{figure*}[t]
\centering
\setlength{\abovecaptionskip}{1mm}
\includegraphics[width=0.85\linewidth]{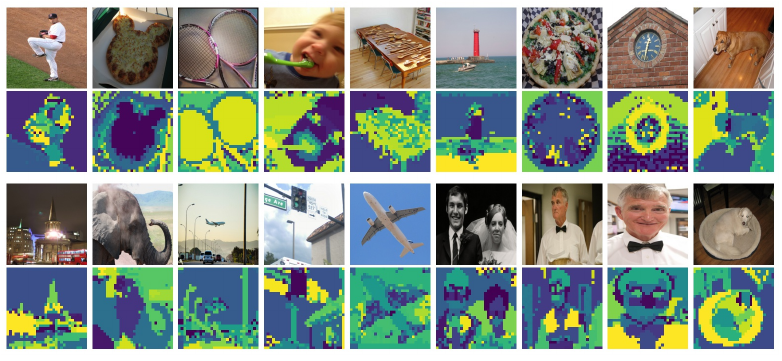}
\caption{Visualization of pattern layout of full images. Different patterns are shown in different colors.}
\label{fig: pattern_layout}
\end{figure*}

\vspace{0.3em}
\noindent\textbf{The effect of the mask ratio in MIM.}
In Table~\ref{tab:mask ratio}, we experiment with different mask ratios for MIM. 
The best performance is obtained at 30\% mask ratio. We analyze that a smaller mask ratio makes some masked patches can be recovered by surrounding patches without utilizing text information, impairing fine-grained cross-modal alignment.
In contrast, too high mask ratio may cause severe information loss, leading to a performance drop.

\vspace{0.3em}
\noindent\textbf{The effect of the start visual layer used to inject textual information for MIM.}
Table \ref{tab: visual layer} shows that fusing textual information starting from different layers of the image encoder has varying effects.
The first line is MIM without using textual information.
As can be seen, injecting text features into shallow layers causes performance drop. We analyze this is due to high-level textual semantics interfere the modeling of low-level visual information.
Starting from deeper layers brings significant performance improvement.
This shows that the injected textual semantics can help high-level visual features to better recover masked regions, thereby facilitating cross-modal semantic alignment.

\subsection{Visualization}

\noindent\textbf{Pattern Cluster.}
As shown in Figure~\ref{fig: pattern_cluster}, we visualize the pattern clusters of the image patch encodings obtained by our model and dVAE~\cite{DBLP:conf/icml/RameshPGGVRCS21}.
For case(a), we can find that the patches are grouped with certain semantic meanings, \textit{e.g.,} \textit{hand} despite their complicated background. 
This demonstrates that our proposed local semantics enhancing approach can effectively inject high-level semantics into local patch encodings, thus providing more semantically meaningful supervision for cross-modal MIM.
In contrast, as case (b) shows, the patches grouped by dVAE share similar textures or colors but have little in common in semantics.
This is because dVAE is trained through image reconstruction which simply minimizes pixel-level differences between original and reconstructed images.
Therefore, it concentrates more on relatively low-level visual information rather than high-level semantics.

\vspace{0.3em}
\noindent\textbf{Pattern Layout of Full Images.}
In Figure~\ref{fig: pattern_layout}, we visualize all the patterns of the image patch encodings in full images, where different patterns are distinguished by different colors. From the outline of the patterns, we can find that the extracted patterns are discriminative. 
This shows that, benefiting from the proposed local semantics enhancing approach, our model can effectively obtain high-level semantic information in image patches.
Therefore, our SemMIM framework can provide MIM with more semantically meaningful reconstruction targets, thereby more effectively promoting cross-modal semantic alignment.

\begin{table}[t]
\setlength\tabcolsep{4pt}
\centering
\scalebox{0.65}{
\begin{tabular}{l|c|c|cc}
\toprule[1.0pt]
\multicolumn{1}{l|}{\multirow{2}{*}{Methods}}      &
\multicolumn{1}{l|}{\multirow{2}{*}{\#Param}}      &
\multicolumn{1}{c|}{Pre-train Time} &
\multicolumn{2}{c}{MSCOCO} \\
\multicolumn{1}{l|}{ } & \multicolumn{1}{l|}{ } & \multicolumn{1}{c|}{(GPU Days))} & TR@1 & IR@1 \\

\midrule   
ALBEF \cite{li2021align} & 210M & 28 (A100) & 73.1 & 56.8 \\
FLM$_{large}$ \cite{wang2023accelerating} & - & 57 (V100) & 73.5 & 56.6 \\
VinVL \cite{zhang2021vinvl} & - & 320 (V100) & 74.6 & 58.1 \\
METER \cite{dou2022empirical} & - & 64 (A100) & 76.2 & 57.1 \\
BLIP \cite{li2022blip}  & 446M & 112 (A100) & 80.6 & 63.1 \\
mPLUG \cite{li2022mplug} & 354M & 40 (A100) & 80.5 & 63.3 \\
\midrule
Ours & 360M & 32 (A100) & \textbf{81.5} & \textbf{64.2} \\

\bottomrule[1.0pt]
\end{tabular}}
\caption{Comparison of the number of trainable parameters and the pre-training time.} 
\label{table:efficiency}
\end{table}

\subsection{Efficiency}
\noindent \textbf{The number of parameters.}
Table \ref{table:efficiency} shows the comparison of the number of trainable parameters and the pre-training time with some existing methods.
ALBEF \cite{li2021align} has fewer parameters than ours, mainly because it hasn't decoder and thus hasn't caption generation ability. Compared to mPLUG \cite{li2022mplug}, with a mere 1.7\% increase in the number of parameters, our method achieves significantly better performance.
The reason why our method doesn't substantially increase the number of parameters is that, the parameters of the momentum model are acquired via Exponential Moving Average (EMA), rather than trainable parameters. Therefore, our method in fact only adds the parameters of the encoding head, an MLP module with less than 6M parameters. Besides, our masking strategy and injecting textual information do not add any learnable parameters.

\vspace{0.3em}
\noindent \textbf{Training efficiency.}
As Table \ref{table:efficiency} shows, compared to existing methods, our method requires almost the shortest pre-training time
(except for ALBEF), while achieving state-of-the-art performance across various vision-language tasks.
On one hand, compared to the baseline, most operations we add are computationally efficient. For example, the momentum model only needs a forward pass to extract image features without requiring back-propagation. Injecting textual information only increases the sequence length slightly, as the text length is much shorter than the number of image patches. On the other hand, our SemMIM framework effectively promotes cross-modal semantic alignment, and thus accelerates the convergence of training, requiring fewer training epochs.

\vspace{0.3em}
\noindent \textbf{Inference efficiency.}
It's also worth noting that, although we add the cross-modal MIM task during pre-training, this only affects the pre-training time and has no impact on the inference speed. After training, our model maintains similar inference efficiency to other models with comparable size and architecture such as mPLUG \cite{li2022mplug}.

\section{Conclusion}
In this work, we have proposed a semantics-enhanced cross-modal MIM framework (SemMIM) for fine-grained vision-language semantic alignment.
Firstly, we propose to inject high-level semantics into local patch encodings via self-supervised agreement learning and sharing encoding space, which can provide more semantically meaningful supervision for MIM.
And we propose a text-guided masking strategy and devise an efficient way of injecting textual information in both masked modeling and reconstruction target acquisition, which achieves deep involvement of text during the entire MIM process.
Extensive experiments on various vision-language tasks and the visualization results verify the effectiveness of the proposed SemMIM method.

\section{Acknowledgements}
This work is supported by the National Key Research and Development Program of China (Grant No. 2022ZD0118501), Beijing Natural Science Foundation (Grant No. JQ21017, L223003, 4224093), the Natural Science Foundation of China (Grant No. 61972397, 62036011, 62192782, U2033210, 62202470), The Project of Beijing Science and technology Committee (Project No. Z231100005923046).

\section{Bibliographical References}\label{sec:reference}

\bibliographystyle{lrec-coling2024-natbib}
\bibliography{lrec-coling2024-example}


\end{document}